\documentclass[10pt,letterpaper]{article}
\usepackage[letterpaper,margin=1in]{geometry}
\usepackage[T1]{fontenc}
\usepackage[utf8]{inputenc}
\usepackage{amsmath,amsthm,mathtools}
\usepackage{amssymb}
\usepackage{newtxtext,newtxmath}
\usepackage{graphicx,booktabs,array,tabularx}
\usepackage[table,dvipsnames]{xcolor}
\usepackage{enumitem,microtype}
\usepackage[font=small,labelfont=bf]{caption}
\usepackage{url}
\usepackage[colorlinks=true,citecolor=blue,linkcolor=blue,urlcolor=blue]{hyperref}
\usepackage{bookmark}

\newtheorem{theorem}{Theorem}
\newtheorem{proposition}{Proposition}

\theoremstyle{remark}
\newtheorem{remark}{Remark}

\theoremstyle{definition}

\newcommand{\R}{\mathbb{R}}

\newcommand{\SO}{\mathrm{SO}}
\newcommand{\so}{\mathfrak{so}}
\newcommand{\su}{\mathfrak{su}}

\DeclareMathOperator{\diag}{diag}

\DeclareMathOperator{\Proj}{Proj}
\DeclareMathOperator{\softmax}{softmax}

\setlist{nosep,topsep=4pt}
\graphicspath{{figs/}{./}}

\hypersetup{
    colorlinks=true,
    linkcolor=blue,
    citecolor=blue,
    urlcolor=blue
}
\usepackage[most]{tcolorbox}

\definecolor{skyblue}{RGB}{100, 186, 255}   

\newtcolorbox{newabstract}{
    colback=skyblue!12!white, 
    colframe=skyblue!6!white,   
    boxrule=0pt,             
    arc=1mm,                 
    left=3pt, right=3pt,   
    top=5pt, bottom=5pt,   
    width=\textwidth,        
}

\newtcolorbox{eqbox}{
  colback=skyblue!6!white,     
  colframe=gray!30,     
  boxrule=0pt,           
  arc=2pt,               
  left=6pt,right=6pt,    
  top=4pt,bottom=4pt,    
  boxsep=0pt,
  enhanced,
  breakable              
}

\title{\textbf{GaugeAttention: Holonomic Self-Attention in Curved Spaces}}
\author{Zhongping Ji\\
\\
jzp@hdu.edu.cn\\
Hangzhou Dianzi University}
\date{}

\begin{document}

\maketitle

\begin{newabstract}
We introduce \textbf{GaugeAttention}, a self-attention mechanism that embeds tokens on a curved manifold and compares queries and keys by directly learning the parallel transport mappings between their distinct tangent spaces, without explicitly constructing the underlying continuous paths. Each token is associated with a manifold coordinate and a local tangent space. The inter-tangent transport operator $\mathcal{P}_{n\to m}=\mathbf{T}_m\exp(\mathbf{\Omega}_{nm})\mathbf{T}_n^{-1}$ inserts a Lie-algebraic curvature rotation between local frames within a common reference space. Composing these transport operators around a closed loop captures the intrinsic holonomy of the curved manifold. Using $\so(4)\cong\su(2)_L\oplus\su(2)_R$, we factor each four-dimensional block into commuting rotations for flat relative displacement and an intrinsic curvature phase. The spatial wedge phase yields translation-invariant loop holonomy, while the content-dependent phase preserves scores under coordinate shifts with features held fixed. A mean-normalized conformal fiber metric supplies the separable gate $s_m s_n$ outside softmax, scaling source values and destination outputs. The identity $\mathbf{J}_R^2=-\mathbf{I}$ enables exact evaluation of non-separable phases in $\mathcal{O}(N^2D)$ time via two matrix multiplications and elementwise modulation. Separable rotations can be evaluated exactly with FlashAttention, and a weak-field feature expansion supports first-order evaluation of wedge phases. A separate linear variant aggregates kernelized features in $\mathcal{O}(ND^2)$ time. On CIFAR-100, the best variant reaches a three-seed mean top-1 accuracy of $69.61\%$, $4.10$ percentage points above the matched ViT baseline with learned absolute positional embeddings.
\end{newabstract}


 \section{Introduction}

Self-attention lets each token gather information from other tokens. It is a central part of Transformers~\cite{vaswani2017attention} used in language models~\cite{devlin2019bert,raffel2020exploring} and vision models~\cite{dosovitskiy2021image,touvron2021training,carion2020end}. Given $N$ tokens, each attention head projects them into queries $\mathbf{Q}$, keys $\mathbf{K}$, and values $\mathbf{V}$. Query--key dot products determine the weights used to combine the values.

\begin{eqbox}
\begin{equation}
    \mathrm{Attention}(\mathbf{Q}, \mathbf{K}, \mathbf{V}) = \softmax\left(\frac{\mathbf{Q}\mathbf{K}^\top}{\sqrt{d_k}}\right)\mathbf{V}.
    \label{eq:vanilla_attn}
\end{equation}
\end{eqbox}

Equation~\eqref{eq:vanilla_attn} uses token features to compute the weights. Without positional information, reordering the tokens simply reorders the outputs. In vision, spatial structure matters because patches with similar features can occupy very different positions within an object. Vision Transformers therefore rely on architectural choices, such as convolutional stems~\cite{dosovitskiy2021image,xiao2021early} or local attention windows, to provide spatial cues~\cite{liu2021swin,bronstein2021geometric}. Softmax also makes every query distribute a total weight of one. The distribution can concentrate on a few tokens, as observed in attention sinks~\cite{xiao2024efficient}. Adding a geometric term for every query--key pair can introduce an $N\times N$ intermediate, making it harder to use memory-efficient kernels such as FlashAttention~\cite{dao2022flashattention} or to aggregate tokens in linear time~\cite{katharopoulos2020transformers,yang2024gated,gu2024mamba}.

Absolute positional encodings add fixed sinusoidal~\cite{vaswani2017attention} or learned~\cite{devlin2019bert,dosovitskiy2021image} vectors to token features. Relative biases modify attention scores using information about token offsets~\cite{dai2019transformer,raffel2020exploring,ke2021rethinking,press2022train}, and conditional convolutions use local neighborhoods~\cite{chu2023conditional,xiao2021early}. Rotary position embedding (RoPE) rotates queries and keys so that their inner product depends on relative position~\cite{su2024roformer}. This approach has been extended to longer sequences and two-dimensional image grids~\cite{sun2023length,heo2024rotary}. DAPE~\cite{zheng2024dape} and CoPE~\cite{golovneva2024contextual} make positional information depend on token content.

The idea of local coordinate frames gives a useful way to interpret rotary encodings~\cite{bronstein2021geometric,cohen2019gauge}. Assign a frame $\mathbf{T}_i$ to each token $i$. A rotation such as RoPE~\cite{su2024roformer} then compares tokens at $n$ and $m$ using a map of the form $\mathcal{P}_{n\to m}=\mathbf{T}_m\mathbf{T}_n^{-1}$. Along the triangle $n\to m\to l\to n$, the intermediate frames cancel.
\begin{equation*}
    \mathcal{P}_{n\to m\to l\to n}^{\mathrm{flat}}
    = (\mathbf{T}_n\mathbf{T}_l^{-1})(\mathbf{T}_l\mathbf{T}_m^{-1})(\mathbf{T}_m\mathbf{T}_n^{-1})
    = \mathbf{I}.
\end{equation*}
The transported vector returns unchanged. In gauge theory, the net transformation around a closed loop is called \emph{holonomy}~\cite{yang1954conservation,cohen2019gauge}. A non-identity result can encode information such as the signed area inside a loop. Content-dependent methods can introduce a different kind of dependence between tokens. For instance, CoPE determines relative positions by accumulating contributions along a sequence~\cite{golovneva2024contextual}.

We also want token content to control how strongly features are passed between tokens. Multiplying all logits for a query by a positive scalar changes the sharpness of its softmax distribution, whose weights still sum to one. A gate applied after softmax can scale the output directly. In linear attention, a source gate can similarly control how much each token contributes to the shared memory~\cite{katharopoulos2020transformers}.

We introduce \textbf{GaugeAttention}. Following the local frame construction of gauge theory~\cite{yang1954conservation}, we insert a rotation $\exp(\mathbf{\Omega}_{nm})$ between the two frame maps. The resulting transport $\mathcal{P}_{n\to m}=\mathbf{T}_m\exp(\mathbf{\Omega}_{nm})\mathbf{T}_n^{-1}$ moves the key at $n$ into the query's frame at $m$. The rotation can depend on the two token positions or their content. Values are aggregated separately, with a learned scale at both the source and the destination.

\begin{itemize}[leftmargin=1.5em]
    \item \textbf{Transport with non-trivial holonomy.} We use two commuting rotations in each four-dimensional block. One carries relative displacement, and the other carries a pairwise phase. With the spatial wedge phase $x_n y_m-y_n x_m$, the phase accumulated around a triangle is proportional to its signed area. A second formulation learns the phase from token features.
    \item \textbf{Exact computation and two efficient variants.} The identity $\mathbf{J}_R^2=-\mathbf{I}$ gives an exact $\mathcal{O}(N^2D)$ score computation using two matrix multiplications and elementwise operations (Theorem~\ref{thm:fast_holonomy}). Separable rotations and a first-order feature expansion for wedge phases give two ways to use FlashAttention. A separate linear attention variant aggregates values into a shared memory in $\mathcal{O}(ND^2)$ time.
    \item \textbf{Content-dependent metric gating.} The pairwise gate $s_m s_n$ applies the key scale to the values before aggregation and the query scale to the output afterward. The scales are normalized by their mean. An optional spatial weighting step, Locality Focusing, controls the contribution of nearby tokens.
\end{itemize}

On CIFAR-100, the content-dependent wedge variant with dynamic metric gating and Locality Focusing reaches $69.61\%$ mean top-1 accuracy over seeds $42$, $43$, and $44$. The matched Conv-Stem ViT with learned absolute positional embeddings reaches $65.51\%$, a difference of $4.10$ percentage points.

\section{The Attention Manifold and Gauge Parallel Transport}

Consider a sequence of $N$ input tokens with embeddings $\{\mathbf{x}_i\}_{i=1}^N$, where $\mathbf{x}_i \in \R^D$ is devoid of explicit positional encoding. We posit that these tokens reside at discrete points $\{\mathbf{p}_i\}_{i=1}^N$ (e.g., 1D sequence indices or 2D spatial grid coordinates $\mathbf{p}_{ij}$) on a curved manifold $\mathcal{M}$, which we term the \textbf{Attention Manifold}.

\subsection{Fiber Bundle Formulation of Queries, Keys, and Values}

In differential geometry, the collection of all tangent spaces $\{T_{\mathbf{p}}\mathcal{M}\}_{\mathbf{p}\in\mathcal{M}}$ forms the \textbf{Tangent Bundle} $T\mathcal{M}$. We conceptualize the query and key representations as sections of $T\mathcal{M}$: at each point $\mathbf{p}_i \in \mathcal{M}$, the model samples tangent vectors $\mathbf{q}_i, \mathbf{k}_i \in T_{\mathbf{p}_i}\mathcal{M}$ of dimension $D$. Each tangent space $T_{\mathbf{p}_m}\mathcal{M}$ is equipped with a local Riemannian metric tensor $\mathbf{M}_m \succ 0$, which defines the local inner product $\langle \mathbf{u}, \mathbf{v} \rangle_{\mathbf{M}_m} = \mathbf{u}^\top \mathbf{M}_m \mathbf{v}$ for $\mathbf{u}, \mathbf{v} \in T_{\mathbf{p}_m}\mathcal{M}$.

Because a query $\mathbf{q}_m \in T_{\mathbf{p}_m}\mathcal{M}$ and a key $\mathbf{k}_n \in T_{\mathbf{p}_n}\mathcal{M}$ ($m \neq n$) inhabit distinct tangent spaces, their inner product $\mathbf{q}_m^\top \mathbf{M}_m \mathbf{k}_n$ cannot be computed directly. A parallel transport operator $\mathcal{P}_{n\to m}: T_{\mathbf{p}_n}\mathcal{M} \to T_{\mathbf{p}_m}\mathcal{M}$ is required to move $\mathbf{k}_n$ along a curve connecting $\mathbf{p}_n$ and $\mathbf{p}_m$ into $T_{\mathbf{p}_m}\mathcal{M}$, yielding the geometrically well-defined similarity score:

\begin{eqbox}
\begin{equation}
    \mathrm{Score}(\mathbf{q}_m, \mathbf{k}_n) = \langle \mathbf{q}_m, \mathcal{P}_{n\to m}(\mathbf{k}_n) \rangle_{\mathbf{M}_m} = \mathbf{q}_m^\top \mathbf{M}_m \mathcal{P}_{n\to m}(\mathbf{k}_n).
\end{equation}
\end{eqbox}

In stark contrast, we formulate the value representations $\{\mathbf{v}_i\}_{i=1}^N$ as sections of a separate \textbf{Trivial Vector Bundle} $\mathcal{E}_V = \mathcal{M} \times \R^{D_v}$ (which can subsequently be endowed with its own independent gauge connection, as explored in Section~\ref{sec:value_bundle}). This architectural decoupling offers two major advantages: first, it separates the intrinsic geometric alignment of $Q/K$ (determining \emph{where to attend}) from the feature aggregation of $V$ (determining \emph{what information to integrate}); second, the fiber dimension $D_v$ does not need to match the tangent space dimension $D$.

\subsection{Foundational Transport and the Limitation of Flat Connections}

In continuous differential geometry, parallel transport of a vector $v^k$ along a curve $c(t)$ is governed by the vanishing covariant derivative $\frac{dv^k}{dt} + \Gamma^k_{ij}\frac{dc^i}{dt}v^j = 0$, where $\Gamma^k_{ij}$ are the Christoffel symbols. Because explicitly parameterizing continuous curves and Christoffel symbols in a discrete, high-dimensional neural network is computationally intractable, we instead adopt a gauge-theoretic frame bundle approach.

Let $\R^D$ serve as a common local reference space (the standard fiber). At each point $\mathbf{p}_i \in \mathcal{M}$, we define a linear frame map (or \emph{vierbein} / local gauge choice) $\mathcal{P}_i: \R^D \to T_{\mathbf{p}_i}\mathcal{M}$, represented by an invertible $D \times D$ matrix $\mathbf{T}_i$. Its inverse $\mathbf{T}_i^{-1}: T_{\mathbf{p}_i}\mathcal{M} \to \R^D$ acts as a \emph{tangent aligning} map that projects local tangent vectors back into the common reference frame.

\paragraph{The Foundational (Flat) Operator.} A natural first-step construction for transporting $\mathbf{k}_n \in T_{\mathbf{p}_n}\mathcal{M}$ to $T_{\mathbf{p}_m}\mathcal{M}$ is to project $\mathbf{k}_n$ to the reference space $\R^D$ via $\mathbf{T}_n^{-1}$ and then map it into $T_{\mathbf{p}_m}\mathcal{M}$ via $\mathbf{T}_m$:

\begin{eqbox}
\begin{equation}
    \mathcal{P}_{n\to m}^{\mathrm{flat}}(\mathbf{k}_n) = (\mathcal{P}_m \circ \mathcal{P}_n^{-1})(\mathbf{k}_n) = \mathbf{T}_m \mathbf{T}_n^{-1} \mathbf{k}_n.
    \label{eq:flat_transport}
\end{equation}
\end{eqbox}

For $\mathcal{P}_{n\to m}^{\mathrm{flat}}$ to be an isometry that preserves vector magnitudes under the metric—i.e., $\|\mathbf{k}_n\|_{\mathbf{M}_n}^2 = \|\mathcal{P}_{n\to m}^{\mathrm{flat}}(\mathbf{k}_n)\|_{\mathbf{M}_m}^2$—it must satisfy:

\begin{eqbox}
\begin{equation}
    \mathbf{k}_n^\top \mathbf{M}_n \mathbf{k}_n = \mathbf{k}_n^\top \left(\mathbf{T}_n^{-1}\right)^\top \mathbf{T}_m^\top \mathbf{M}_m \mathbf{T}_m \mathbf{T}_n^{-1} \mathbf{k}_n \implies \mathbf{T}_n^\top \mathbf{M}_n \mathbf{T}_n = \mathbf{T}_m^\top \mathbf{M}_m \mathbf{T}_m = \mathbf{I}.
    \label{eq:metric_compatibility}
\end{equation}
\end{eqbox}

Decomposing $\mathbf{T}_i = \mathbf{S}_i \mathbf{R}_i$ into a symmetric positive-definite scaling matrix $\mathbf{S}_i = \mathbf{M}_i^{-1/2}$ and an orthogonal rotation matrix $\mathbf{R}_i \in \SO(D)$, Eq.~\eqref{eq:metric_compatibility} is identically satisfied, and the metric inner product reduces to $\langle \mathbf{T}_m^{-1}\mathbf{q}_m, \mathbf{T}_n^{-1}\mathbf{k}_n \rangle_{\mathbf{I}}$.

\begin{remark}[Path-Independence Implies Zero Curvature]
While Eq.~\eqref{eq:flat_transport} elegantly aligns tangent spaces, it exhibits strict transitivity:

\begin{eqbox}
\begin{equation}
    \mathcal{P}_{n\to m \to l}^{\mathrm{flat}} = (\mathbf{T}_l \mathbf{T}_m^{-1})(\mathbf{T}_m \mathbf{T}_n^{-1}) = \mathbf{T}_l \mathbf{T}_n^{-1} = \mathcal{P}_{n\to l}^{\mathrm{flat}}.
\end{equation}
\end{eqbox}

Consequently, transporting any vector around a closed loop $n \to m \to l \to n$ yields the identity operator $\mathcal{P}_{\text{loop}}^{\mathrm{flat}} = \mathbf{I}$. In gauge theory, this means the connection is \emph{pure gauge} (flat), and its Ambrose-Singer holonomy group is trivial. It cannot capture path-dependent geometric phases or intrinsic manifold curvature.
\end{remark}

\subsection{New Paradigm from Gauge Field Theory: The Holonomic ``Sandwich'' Protocol}

\begin{figure}[t!]
    \centering
    \includegraphics[width=\textwidth]{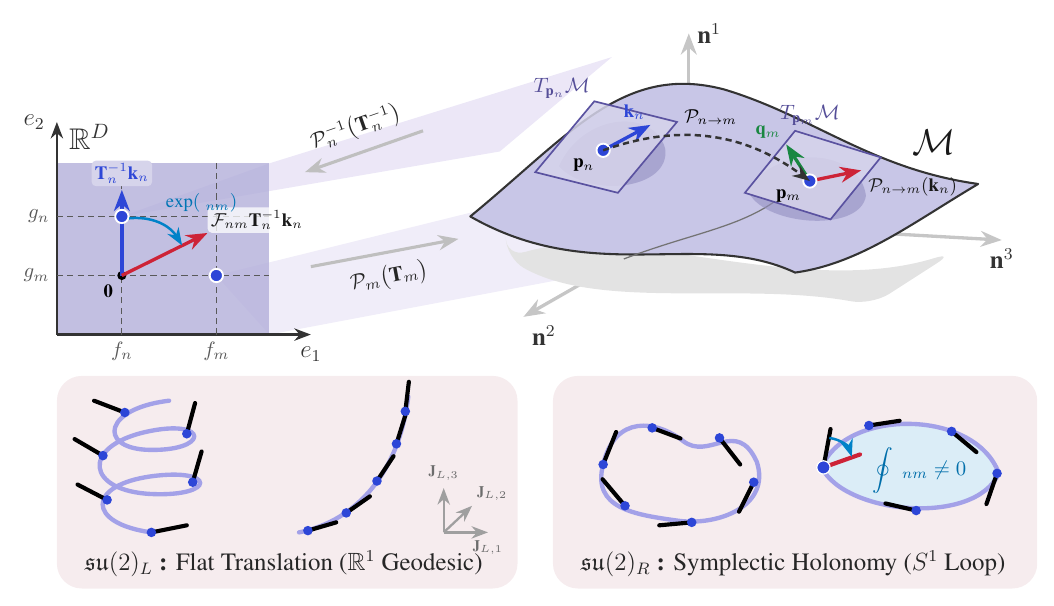}
    \caption{\textbf{Top:} Geometric architecture of the Gauge ``Sandwich'' Parallel Transport protocol $\mathcal{P}_{n\to m} = \mathbf{T}_m \cdot \exp(\mathbf{\Omega}_{nm}) \cdot \mathbf{T}_n^{-1}$. Tangent vectors $\mathbf{k}_n \in T_{\mathbf{p}_n}\mathcal{M}$ are aligned into the common reference fiber $\R^D$ via $\mathcal{P}_n^{-1}$, rotated by the Lie-algebraic curvature operator $\exp(\mathbf{\Omega}_{nm})$, and lifted to $T_{\mathbf{p}_m}\mathcal{M}$ via $\mathcal{P}_m$. \textbf{Bottom:} Chiral $\so(4) \cong \su(2)_L \oplus \su(2)_R$ trajectories, where $\su(2)_L$ governs path-independent geodesic translation (left) and $\su(2)_R$ captures non-trivial closed-loop holonomy $\oint \mathbf{\Omega}_{nm} \neq 0$ (right).}
    \label{fig:gauge_manifold}
\end{figure}

Adapting the Transformer to higher-dimensional structured data—such as 2D images, 3D voxels, or complex semantic graphs—requires modeling non-trivial curvature and topological holonomy. To break the assumption of spatial flatness, we introduce an \textbf{interaction operator} (discrete Wilson line / gauge transporter) $\mathcal{F}_{nm}: \R^D \to \R^D$ acting within the common reference space $\R^D$. This generalizes our parallel transport protocol into a gauge-theoretic \textbf{``sandwich'' structure}:

\begin{eqbox}
\begin{equation}
    \mathcal{P}_{n\to m} = \mathbf{T}_m \cdot \mathcal{F}_{nm} \cdot \mathbf{T}_n^{-1}.
    \label{eq:sandwich_protocol}
\end{equation}
\end{eqbox}
In this formulation, $\mathbf{T}_n^{-1}$ maps the key vector from $T_{\mathbf{p}_n}\mathcal{M}$ to the reference fiber $\R^D$, the intermediate term $\mathcal{F}_{nm}$ applies a path- and interaction-dependent transformation encoding the manifold's intrinsic geometry, and $\mathbf{T}_m$ lifts the result into the destination tangent space $T_{\mathbf{p}_m}\mathcal{M}$ (see Fig.~\ref{fig:gauge_manifold}).

To preserve the norm of feature vectors during parallel transport (satisfying the metric compatibility condition in Eq.~\eqref{eq:metric_compatibility}) and focus on directional geometric relationships, we instantiate $\mathcal{F}_{nm}$ as an orthogonal rotation within the structure group $\SO(D)$. By Lie's theorem, this transformation is generated via the matrix exponential of an element in the skew-symmetric Lie algebra $\so(D)$, denoted as the curvature/holonomy matrix $\mathbf{\Omega}_{nm} \in \so(D)$:

\begin{eqbox}
\begin{equation}
    \mathcal{F}_{nm} = \exp(\mathbf{\Omega}_{nm}), \quad \text{where} \quad \mathbf{\Omega}_{nm}^\top = -\mathbf{\Omega}_{nm}.
    \label{eq:lie_generator}
\end{equation}
\end{eqbox}
Substituting Eq.~\eqref{eq:lie_generator} into Eq.~\eqref{eq:sandwich_protocol} yields our holonomic parallel transport operator:

\begin{eqbox}
\begin{equation}
    \mathcal{P}_{n\to m} = \mathbf{T}_m \cdot \exp(\mathbf{\Omega}_{nm}) \cdot \mathbf{T}_n^{-1}.
    \label{eq:enhanced_transport}
\end{equation}
\end{eqbox}
Transporting a vector around the triangle $(n,m,l)$ yields $\mathcal{P}_{\mathrm{loop}}=\mathbf{T}_n\exp(\mathbf{\Omega}_{ln})\exp(\mathbf{\Omega}_{ml})\exp(\mathbf{\Omega}_{nm})\mathbf{T}_n^{-1}$. A non-identity loop operator encodes non-trivial holonomy. For the fixed-axis construction below, this condition becomes $\lambda(\mathit{\Omega}_{nm}+\mathit{\Omega}_{ml}+\mathit{\Omega}_{ln})\notin 2\pi\mathbb{Z}$.

\section{Chiral $\SO(4)$ Decomposition and Curvature Instantiations}
\label{sec:so4_decomposition}

A general skew-symmetric matrix in $\so(D)$ possesses $D(D-1)/2$ independent parameters, and computing its matrix exponential $\exp(\mathbf{\Omega}_{nm})$ for all $N \times N$ token pairs is computationally prohibitive. To achieve a parameter-efficient, closed-form, and physically grounded realization of Eq.~\eqref{eq:enhanced_transport}, we systematically decompose high-dimensional rotations using Lie group theory. We designate our family of gauge-transport schemes as \textbf{Gauge-$\SO(d)$}.

\subsection{From Abelian $\SO(2)$ and $\SO(3)$ to Chiral $\SO(4)$ Subgroups}

\paragraph{Commuting $\SO(2)$ Subgroups (Maximal Torus, $\mathrm{Gauge\text{-}SO(2)}$).}
For $D = 2k$, the Cartan subalgebra $\mathfrak{h} = \bigoplus_{i=1}^k \so(2)^{(i)} \subset \so(2k)$ consists of block-diagonal matrices $H = \diag(\theta_1 J, \dots, \theta_k J)$, where $J = \begin{bsmallmatrix} 0 & -1 \\ 1 & 0 \end{bsmallmatrix}$ satisfies $J^2 = -I_2$. Setting $\mathbf{T}_n = \exp(n H)$ and $\mathbf{\Omega}_{nm} = \mathbf{0}$ yields $\mathcal{P}_{n\to m} = \diag_{i=1}^k\big(\cos((m-n)\theta_i)I_2 + \sin((m-n)\theta_i)J\big)$. When $\mathbf{M}_i = \mathbf{I}$, this reduces to standard RoPE~\cite{su2024roformer}, confirming that RoPE is a special case of GaugeAttention under a flat, abelian $\SO(2)^k$ connection.

\paragraph{Non-Abelian $\SO(3)$ Subgroups.}
For $D = 3k$, rotations in each 3D block are generated by a unit skew-symmetric matrix $K = \begin{bsmallmatrix} 0 & -c & b \\ c & 0 & -a \\ -b & a & 0 \end{bsmallmatrix} \in \so(3)$ with $a^2+b^2+c^2=1$, admitting Rodrigues' rotation formula $\exp(\theta K) = I_3 + \sin(\theta)K + (1-\cos(\theta))K^2$. However, $\so(3)$ is simple and cannot be factored into commuting ideals.

\paragraph{The Exceptional Isoclinic Decomposition of $\SO(4)$ ($\mathrm{Gauge\text{-}SO(4)}$).}
Our core algebraic construction leverages a unique structural property of four-dimensional space: $\so(4)$ is the only semi-simple orthogonal Lie algebra that decomposes into a direct sum of two commuting ideals:

\begin{eqbox}
\begin{equation}
    \so(4) \cong \su(2)_L \oplus \su(2)_R \cong \so(3)_L \oplus \so(3)_R.
\end{equation}
\end{eqbox}
Specifically, the six-dimensional vector space of $4 \times 4$ real skew-symmetric matrices splits into two mutually commuting three-dimensional subalgebras, $\mathfrak{h}_L = \mathrm{span}\{A_1, A_2, A_3\}$ (generating \emph{left-isoclinic} rotations) and $\mathfrak{h}_R = \mathrm{span}\{B_1, B_2, B_3\}$ (generating \emph{right-isoclinic} rotations):

\begin{eqbox}
\begin{equation}
    A_1 = \begin{bmatrix} 0 & 0 & 0 & -1 \\ 0 & 0 & -1 & 0 \\ 0 & 1 & 0 & 0 \\ 1 & 0 & 0 & 0 \end{bmatrix}, \quad
    A_2 = \begin{bmatrix} 0 & 0 & 1 & 0 \\ 0 & 0 & 0 & -1 \\ -1 & 0 & 0 & 0 \\ 0 & 1 & 0 & 0 \end{bmatrix}, \quad
    A_3 = \begin{bmatrix} 0 & -1 & 0 & 0 \\ 1 & 0 & 0 & 0 \\ 0 & 0 & 0 & -1 \\ 0 & 0 & 1 & 0 \end{bmatrix},
\end{equation}
\end{eqbox}
\begin{eqbox}
\begin{equation}
    B_1 = \begin{bmatrix} 0 & 0 & 0 & 1 \\ 0 & 0 & -1 & 0 \\ 0 & 1 & 0 & 0 \\ -1 & 0 & 0 & 0 \end{bmatrix}, \quad
    B_2 = \begin{bmatrix} 0 & 0 & 1 & 0 \\ 0 & 0 & 0 & 1 \\ -1 & 0 & 0 & 0 \\ 0 & -1 & 0 & 0 \end{bmatrix}, \quad
    B_3 = \begin{bmatrix} 0 & -1 & 0 & 0 \\ 1 & 0 & 0 & 0 \\ 0 & 0 & 0 & 1 \\ 0 & 0 & -1 & 0 \end{bmatrix}.
\end{equation}
\end{eqbox}
Given any unit axis vectors $\mathbf{a} = (a_1, a_2, a_3)^\top \in S^2$ and $\mathbf{b} = (b_1, b_2, b_3)^\top \in S^2$, we construct the normalized chiral generators $\mathbf{J}_L \in \mathfrak{h}_L$ and $\mathbf{J}_R \in \mathfrak{h}_R$:

\begin{eqbox}
\begin{equation}
    \mathbf{J}_L = \sum_{i=1}^3 a_i A_i = \begin{bmatrix} 0 & -a_3 & a_2 & -a_1 \\ a_3 & 0 & -a_1 & -a_2 \\ -a_2 & a_1 & 0 & -a_3 \\ a_1 & a_2 & a_3 & 0 \end{bmatrix}, \qquad
    \mathbf{J}_R = \sum_{i=1}^3 b_i B_i = \begin{bmatrix} 0 & -b_3 & b_2 & b_1 \\ b_3 & 0 & -b_1 & b_2 \\ -b_2 & b_1 & 0 & b_3 \\ -b_1 & -b_2 & -b_3 & 0 \end{bmatrix}.
\end{equation}
\end{eqbox}

\begin{proposition}[Commutativity and Euler-Like Closed Forms]
\label{prop:so4_properties}
For any unit vectors $\mathbf{a}, \mathbf{b} \in S^2$, the chiral generators $\mathbf{J}_L$ and $\mathbf{J}_R$ satisfy:
\begin{enumerate}[label=(\roman*)]
    \item \textbf{Chiral Commutativity:} $[\mathbf{J}_L, \mathbf{J}_R] = \mathbf{J}_L \mathbf{J}_R - \mathbf{J}_R \mathbf{J}_L = \mathbf{0}$. Consequently, $\exp(\theta_L \mathbf{J}_L + \theta_R \mathbf{J}_R) = \exp(\theta_L \mathbf{J}_L)\exp(\theta_R \mathbf{J}_R)$.
    \item \textbf{Complex-Like Involution:} $\mathbf{J}_L^2 = -\mathbf{I}_4$ and $\mathbf{J}_R^2 = -\mathbf{I}_4$. Thus, their matrix exponentials admit exact Euler-like closed forms:

    \begin{eqbox}
    \begin{equation}
        \exp(\theta_L \mathbf{J}_L) = \cos(\theta_L)\mathbf{I}_4 + \sin(\theta_L)\mathbf{J}_L, \qquad \exp(\theta_R \mathbf{J}_R) = \cos(\theta_R)\mathbf{I}_4 + \sin(\theta_R)\mathbf{J}_R.
        \label{eq:euler_so4}
    \end{equation}
    \end{eqbox}
\end{enumerate}
\end{proposition}
By embedding $k$ copies of $\SO(4)$ into block-diagonal matrices in $\SO(D)$ ($D = 4k$), Proposition~\ref{prop:so4_properties} extends directly to $\R^D$.

\subsection{Factorizing the Sandwich Operator into Flat Translation and Symplectic Curvature}

Proposition~\ref{prop:so4_properties} provides the exact algebraic bridge between our gauge ``sandwich'' protocol (Eq.~\eqref{eq:enhanced_transport}) and a tractable geodesic transport operator. Specifically, we assign the two commuting subalgebras $\mathfrak{h}_L$ and $\mathfrak{h}_R$ to two distinct geometric roles:
\begin{itemize}
    \item \textbf{Left-Chiral Local Frame ($\mathfrak{h}_L$) and the Contraction $\mathbf{p}_n \cdot \mathbf{J}_L$:} We parameterize the rotational component of the local frame $\mathbf{T}_n$ at token coordinate $\mathbf{p}_n \in \R^d$ using the left-isoclinic subalgebra: $\mathbf{R}_n = \exp(\mathbf{p}_n \cdot \mathbf{J}_L)$. Formally, the dot product $\mathbf{p}_n \cdot \mathbf{J}_L$ denotes a directional contraction mapping the coordinate vector $\mathbf{p}_n$ into $\mathfrak{h}_L$. Specifically, let $\boldsymbol{\omega}_L \in \R^d$ denote a learnable frequency projection vector and $\mathbf{J}_L = \sum_{i=1}^3 a_i A_i \in \mathfrak{h}_L$ be a normalized generator ($\|\mathbf{a}\|_2 = 1$); then $\mathbf{p}_n \cdot \mathbf{J}_L \triangleq (\mathbf{p}_n^\top \boldsymbol{\omega}_L)\mathbf{J}_L$ scales the unit generator $\mathbf{J}_L$ by the projected coordinate phase $\theta_{L,n} = \mathbf{p}_n^\top \boldsymbol{\omega}_L \in \R$. (More generally, for 3D coordinates $\mathbf{p}_n \in \R^3$, one may also define the direct Lie-algebraic contraction $\mathbf{p}_n \cdot \vec{\mathbf{A}} = \sum_{i=1}^3 p_{n,i} A_i$, whose square yields $-\|\mathbf{p}_n\|_2^2 \mathbf{I}_4$). Under the single-axis projection $\mathbf{p}_n \cdot \mathbf{J}_L = (\mathbf{p}_n^\top \boldsymbol{\omega}_L)\mathbf{J}_L$, any two positions $\mathbf{p}_m$ and $\mathbf{p}_n$ share the same generator $\mathbf{J}_L$ and therefore commute within $\mathfrak{h}_L$: $[\mathbf{p}_m \cdot \mathbf{J}_L, \, \mathbf{p}_n \cdot \mathbf{J}_L] = \mathbf{0}$.
    
    \item \textbf{Right-Chiral Curvature Interaction ($\mathfrak{h}_R$) and Scalar Phase $\mathit{\Omega}_{nm}$:} To bridge the general Lie-algebraic curvature matrix $\mathbf{\Omega}_{nm} \in \so(D)$ in Eq.~\eqref{eq:enhanced_transport} with a tractable scalar geometric phase, we project $\mathbf{\Omega}_{nm}$ onto the normalized right-isoclinic generator $\mathbf{J}_R \in \mathfrak{h}_R$ via the explicit parameterization:
    \begin{eqbox}
    \begin{equation}
        \mathbf{\Omega}_{nm} = \lambda \cdot \mathit{\Omega}_{nm} \cdot \mathbf{J}_R,
        \label{eq:omega_bridge}
    \end{equation}
    \end{eqbox}
    where $\mathit{\Omega}_{nm} \in \R$ (typeset in italic to distinguish it from the matrix $\mathbf{\Omega}_{nm}$) denotes the scalar curvature form between tokens $n$ and $m$, and $\lambda \in \R$ is a learnable coupling strength. This instantiates the intermediate holonomy operator inside the right-isoclinic subalgebra as $\mathcal{F}_{nm} = \exp(\lambda \cdot \mathit{\Omega}_{nm} \cdot \mathbf{J}_R)$.
\end{itemize}

Substituting these definitions into the pure-rotational sandwich protocol $\mathcal{P}_{n\to m} = \mathbf{R}_m \cdot \mathcal{F}_{nm} \cdot \mathbf{R}_n^{-1}$ and invoking both the inter-chiral commutativity $[\mathbf{J}_L, \mathbf{J}_R] = \mathbf{0}$ from Proposition~\ref{prop:so4_properties}(i) and the intra-axis commutativity $[\mathbf{p}_m \cdot \mathbf{J}_L, \, \mathbf{p}_n \cdot \mathbf{J}_L] = \mathbf{0}$, the right-hand frame $\mathbf{R}_n^{-1} = \exp(-\mathbf{p}_n \cdot \mathbf{J}_L)$ commutes past $\mathcal{F}_{nm}$ and merges additively with $\mathbf{R}_m$:
\begin{eqbox}
\begin{align}
    \mathcal{P}_{n\to m} &= \exp\left(\mathbf{p}_m \cdot \mathbf{J}_L\right) \cdot \exp\left(\lambda \cdot \mathit{\Omega}_{nm} \cdot \mathbf{J}_R\right) \cdot \exp\left(-\mathbf{p}_n \cdot \mathbf{J}_L\right) \nonumber \\
    &= \underbrace{\exp\left(\Delta \mathbf{p}_{nm} \cdot \mathbf{J}_L\right)}_{\text{Flat Translation}} \cdot \underbrace{\exp\left(\lambda \cdot \mathit{\Omega}_{nm} \cdot \mathbf{J}_R\right)}_{\text{Symplectic Curvature}},
    \label{eq:factorized_transport}
\end{align}
\end{eqbox}
where $\Delta \mathbf{p}_{nm} = \mathbf{p}_m - \mathbf{p}_n$ and $\Delta \mathbf{p}_{nm} \cdot \mathbf{J}_L = (\Delta \mathbf{p}_{nm}^\top \boldsymbol{\omega}_L)\mathbf{J}_L$. Equation~\eqref{eq:factorized_transport} is a central result of our framework: it factorizes geodesic parallel transport on the Attention Manifold into a product of a \textbf{flat relative translation component} in $\su(2)_L$ and an \textbf{intrinsic curvature/holonomy component} in $\su(2)_R$. Each four-dimensional block uses a fixed pair of learnable axes $\mathbf{J}_L,\mathbf{J}_R$ shared across token pairs. The resulting transport lies in an abelian $\SO(2)\times\SO(2)$ subgroup embedded in $\SO(4)$: for $\mathbf{H}_e=\theta_e\mathbf{J}_L+\beta_e\mathbf{J}_R$, one has $[\mathbf{H}_e,\mathbf{H}_f]=\mathbf{0}$ for any two edges $e,f$. Non-separable edge phases generate non-trivial closed-loop holonomy within this subgroup.

\subsection{Three Principled Instantiations of the Curvature Form $\mathit{\Omega}_{nm}$}

For 2D visual data with normalized grid coordinates $\mathbf{p}_n = (x_n, y_n) \in [-1, 1]^2$, we associate the left-chiral translation with horizontal displacement $\Delta x_{nm}\theta_L = (x_m - x_n)t\theta_L$ across $t \in \{1, \dots, k\}$ harmonic subspaces. Depending on how the right-chiral phase is parameterized, we obtain three distinct geometric regimes:

\paragraph{1) Orthogonal Spatial Coupling (0th-Order Separable Form, $\mathrm{Ortho}$).}
Setting the right-isoclinic phase to encode vertical displacement $\mathit{\Omega}_{nm}^{\mathrm{ortho}} = (y_m - y_n)t\theta_B$ treats the 2D grid as a fibered product where motion along the $y$-axis acts as an external gauge phase relative to the $x$-axis:
\begin{eqbox}
\begin{equation}
    \mathit{\Omega}_{nm}^{\mathrm{ortho}} = (y_m - y_n)t\theta_B \implies \mathcal{P}_{n\to m}^{\mathrm{ortho}} = \exp\big((x_m - x_n)t\theta_L \mathbf{J}_L\big)\exp\big((y_m - y_n)t\theta_B \mathbf{J}_R\big).
    \label{eq:omega_ortho}
\end{equation}
\end{eqbox}
Because Eq.~\eqref{eq:omega_ortho} is additively separable ($f(m) - f(n)$), both left and right rotations can be absorbed directly into individual token representations $\tilde{\mathbf{q}}_m$ and $\tilde{\mathbf{k}}_n$ in $\mathcal{O}(ND)$ time, serving as our base 2D coordinate alignment.

\paragraph{2) Symplectic Area Form / Berry Phase (Non-Separable Magnetic Curvature, $\mathrm{Symplectic}$).}
To superimpose genuine path-dependent holonomy onto the 2D coordinate alignment, we augment the separable vertical phase with the canonical symplectic 2-form $\omega = dx \wedge dy$. Under the symmetric gauge potential $\mathbf{A} = \frac{1}{2}(-y, x)$, transporting a state from $\mathbf{p}_n$ to $\mathbf{p}_m$ accumulates a magnetic flux (or Berry phase) given by the coordinate wedge product (signed area of the triangle formed with the origin):
\begin{eqbox}
\begin{equation}
    \mathit{\Omega}_{nm}^{\mathrm{symp}} = \mathbf{p}_n \wedge \mathbf{p}_m = x_n y_m - y_n x_m.
    \label{eq:omega_symp}
\end{equation}
\end{eqbox}
Crucially, $\mathit{\Omega}_{nm}^{\mathrm{symp}}$ is \emph{non-separable} and breaks path-independence: around any closed triangle $n \to m \to l \to n$, the net holonomy angle is $\mathit{\Omega}_{nm}^{\mathrm{symp}} + \mathit{\Omega}_{ml}^{\mathrm{symp}} + \mathit{\Omega}_{ln}^{\mathrm{symp}} = 2\,\mathrm{Area}(\triangle nml) \neq 0$. This endows the Transformer with an explicit geometric awareness of 2D loop areas, chirality, and rotational orientation.

\paragraph{3) Content-Induced Dynamic Curvature (Abelian Gauge Phase, $\mathrm{Dynamic}$).}
To couple the gauge phase to semantic content, we use token embeddings as the source of a dynamic abelian connection. We parameterize $\mathit{\Omega}_{nm}$ as a bilinear wedge form of the token embeddings $\mathbf{x}_n, \mathbf{x}_m$ via lightweight bounded projections $\boldsymbol{\phi}(\mathbf{x}_i) = \tanh(\mathbf{x}_i \mathbf{W}_\phi) \in \R^{n_h \times 2}$:
\begin{eqbox}
\begin{equation}
    \mathit{\Omega}_{nm}^{\mathrm{dyn}} = \phi_1(\mathbf{x}_n)\phi_2(\mathbf{x}_m) - \phi_2(\mathbf{x}_n)\phi_1(\mathbf{x}_m) \quad \text{or} \quad \mathit{\Omega}_{nm}^{\mathrm{dyn\text{-}sep}} = \phi(\mathbf{x}_m) - \phi(\mathbf{x}_n),
    \label{eq:omega_dyn}
\end{equation}
\end{eqbox}
allowing pairwise semantic interactions to induce data-dependent holonomy deflections in the right-chiral subspace.

\subsection{Exact $\mathcal{O}(N^2 D)$ Evaluation of Non-Separable Curvature via $\mathbf{J}_R^2 = -\mathbf{I}$}

When superimposing non-separable curvature $\mathit{\Omega}_{nm}$ (Eq.~\eqref{eq:omega_symp} or \eqref{eq:omega_dyn}) onto the 2D base coordinate alignment (Eq.~\eqref{eq:omega_ortho}), both terms reside in $\mathfrak{h}_R$ along the same generator $\mathbf{J}_R$ and therefore commute: $\exp(((y_m-y_n)t\theta_B + \lambda\mathit{\Omega}_{nm})\mathbf{J}_R) = \exp((y_m-y_n)t\theta_B\mathbf{J}_R)\exp(\lambda\mathit{\Omega}_{nm}\mathbf{J}_R)$. However, because $n$ and $m$ are multiplicatively coupled inside $\exp(\lambda \mathit{\Omega}_{nm}\mathbf{J}_R)$, naively applying a distinct $D \times D$ rotation matrix to every pair $(m, n)$ would incur a prohibitive $\mathcal{O}(N^2 D^2)$ memory and compute bottleneck.

We now prove that Proposition~\ref{prop:so4_properties}(ii) completely eliminates this bottleneck, reducing the exact evaluation of arbitrary pairwise curvature $\mathit{\Omega}_{nm}$ to two standard $\mathcal{O}(N^2 D)$ matrix multiplications.

\begin{theorem}[Fast Decomposition of Pairwise Symplectic Holonomy]
\label{thm:fast_holonomy}
Let $\tilde{\mathbf{q}}_m = \exp(-y_m t\theta_B \mathbf{J}_R)\exp(-x_m t\theta_L \mathbf{J}_L)\mathbf{q}_m$ and $\tilde{\mathbf{k}}_n = \exp(-y_n t\theta_B \mathbf{J}_R)\exp(-x_n t\theta_L \mathbf{J}_L)\mathbf{k}_n$ denote the query and key vectors pre-aligned by the base 2D chiral rotations ($\mathcal{O}(ND)$ complexity). For any pairwise scalar curvature matrix $\mathbf{\Phi} = [\mathit{\Omega}_{nm}] \in \R^{N \times N}$ shared across a head, the holonomy-transported inner product matrix $\mathbf{A}_{m,n} = \tilde{\mathbf{q}}_m^\top \exp(\lambda \mathit{\Omega}_{nm} \mathbf{J}_R)\tilde{\mathbf{k}}_n$ decomposes exactly as:
\begin{eqbox}
\begin{equation}
    \mathbf{A} = \cos(\lambda \mathbf{\Phi}) \odot \left(\tilde{\mathbf{Q}}\tilde{\mathbf{K}}^\top\right) + \sin(\lambda \mathbf{\Phi}) \odot \left(\tilde{\mathbf{Q}}(\tilde{\mathbf{K}}\mathbf{J}_R^\top)^\top\right),
    \label{eq:fast_holonomy}
\end{equation}
\end{eqbox}
where $\odot$ denotes the element-wise Hadamard product, and $\cos(\cdot), \sin(\cdot)$ are applied element-wise.
\end{theorem}
\begin{proof}
By Proposition~\ref{prop:so4_properties}(ii), $\mathbf{J}_R^2 = -\mathbf{I}$, which yields $\exp(\lambda \mathit{\Omega}_{nm} \mathbf{J}_R) = \cos(\lambda \mathit{\Omega}_{nm})\mathbf{I} + \sin(\lambda \mathit{\Omega}_{nm})\mathbf{J}_R$. Substituting this into the bilinear form gives:
\begin{eqbox}
\begin{equation*}
    \tilde{\mathbf{q}}_m^\top \exp(\lambda \mathit{\Omega}_{nm} \mathbf{J}_R)\tilde{\mathbf{k}}_n = \cos(\lambda \mathit{\Omega}_{nm}) \left(\tilde{\mathbf{q}}_m^\top \tilde{\mathbf{k}}_n\right) + \sin(\lambda \mathit{\Omega}_{nm}) \left(\tilde{\mathbf{q}}_m^\top \mathbf{J}_R \tilde{\mathbf{k}}_n\right).
\end{equation*}
\end{eqbox}
Since $\mathbf{J}_R$ is a fixed $4 \times 4$ block-diagonal skew-symmetric operator independent of $m$, the right-hand vector $\mathbf{J}_R \tilde{\mathbf{k}}_n$ is computed once per token in $\mathcal{O}(ND)$ time, followed by two standard GEMM calls $\tilde{\mathbf{Q}}\tilde{\mathbf{K}}^\top$ and $\tilde{\mathbf{Q}}(\mathbf{J}_R \tilde{\mathbf{K}})^\top$ in $\mathcal{O}(N^2 D)$ time.
\end{proof}

\section{Matter Determines Geometry: Dynamic Metric Gating}
\label{sec:metric_gating}

While the orthogonal gauge transporter $\mathcal{P}_{n\to m} \in \SO(D)$ governs directional alignment and holonomy across tangent spaces, the local geometry within each tangent space $T_{\mathbf{p}_m}\mathcal{M}$ is dictated by the Riemannian metric tensor $\mathbf{M}_m$.

\subsection{Conformal Deformation and Softmax Decoupling}

In general, $\mathbf{M}_m$ can be parameterized as a positive-definite diagonal matrix $\mathbf{M}_m = \diag((s_m^{(1)})^2, \dots, (s_m^{(D)})^2)$. To maximize parameter efficiency, we constrain $\mathbf{M}_m$ to a conformal metric of the form $\mathbf{M}_m = s_m^2 \mathbf{I}$. The associated pairwise metric factor is:
\begin{eqbox}
\begin{equation}
    \mathbf{g}_{mn} = s_m s_n,
\end{equation}
\end{eqbox}
where $s_i > 0$ represents the local metric scale at token $i$. Under our metric compatibility condition $\mathbf{T}_i = s_i^{-1}\mathbf{R}_i$ (with $\mathbf{M}_i = s_i^2 \mathbf{I}$), the metric-induced inner product between $\mathbf{q}_m$ and $\mathcal{P}_{n\to m}(\mathbf{k}_n)$ scales proportionally to $s_m s_n$.

However, directly inserting this scaled inner product inside the standard softmax function, $\softmax\big(\frac{(s_m \tilde{\mathbf{q}}_m)^\top (s_n \tilde{\mathbf{k}}_n)}{\sqrt{d_k}}\big)$, reveals a critical flaw: because softmax normalizes each row ($\sum_n p_{mn} = 1$), the row-wise query scale $s_m$ controls the inverse temperature of the attention distribution, while the total attention weight remains fixed.

To resolve this, we propose the \textbf{GaugeAttention (\textrm{GaugeAttn})} operator, which preserves the probabilistic alignment of softmax while strategically decoupling the pairwise metric factor $\mathbf{S} = [s_m s_n] = \mathbf{L}\mathbf{L}^\top$ outside the softmax as an explicit geometric gating term:
\begin{eqbox}
\begin{equation}
    \mathrm{GaugeAttn}(\mathbf{Q}, \mathbf{K}, \mathbf{V}) = \left( \mathbf{S} \odot \softmax\left( \frac{\mathcal{A}_{\mathrm{gauge}}(\mathbf{Q}, \mathbf{K})}{\sqrt{d_k}} \right) \right) \mathbf{V} = \left( \mathbf{L}\mathbf{L}^\top \odot \softmax\left( \frac{\mathcal{A}_{\mathrm{gauge}}(\mathbf{Q}, \mathbf{K})}{\sqrt{d_k}} \right) \right) \mathbf{V},
    \label{eq:gaugeattn}
\end{equation}
\end{eqbox}
where $\mathcal{A}_{\mathrm{gauge}}(\mathbf{Q}, \mathbf{K})_{m,n} = \langle \mathbf{q}_m, \mathcal{P}_{n\to m}^{\mathrm{rot}}(\mathbf{k}_n) \rangle$ is the pure rotational gauge similarity from Section~\ref{sec:so4_decomposition}, and $\mathbf{L} = [s_1, s_2, \dots, s_N]^\top \in \R^N$ collects the local metric scales.

\subsection{Dynamic Metric Scaling as a ``Gravitational Lens''}

\begin{figure}[th]
    \centering
    \includegraphics[width=\textwidth]{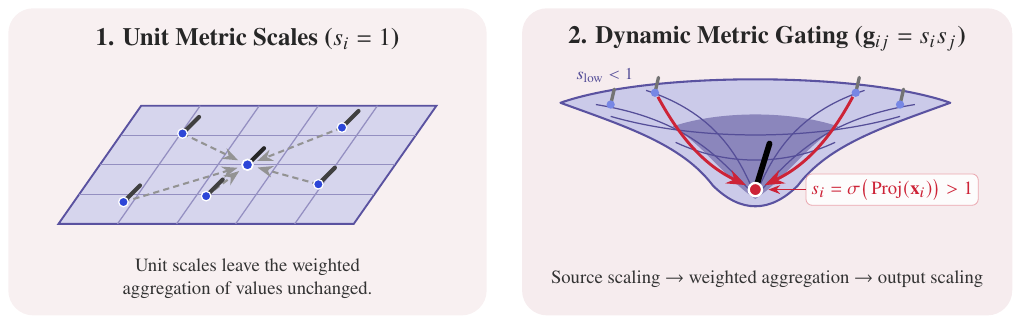}
    \caption{\textbf{Physical mechanism of Metric Gating (Section~\ref{sec:metric_gating}):} Unlike standard attention with a static flat metric $\mathbf{I}$ (left), GaugeAttention learns a content-dependent conformal metric $\mathbf{g}_{ij} = s_i s_j$ decoupled from softmax (right). Salient semantic tokens create ``gravity wells'' that amplify foreground interactions while compressing background noise.}
    \label{fig:gravitational_lens}
\end{figure}

Unlike standard attention where the metric is fixed ($\mathbf{I}$), or static positional schemes where $s_i$ depends solely on spatial coordinates, we draw inspiration from General Relativity: \emph{matter determines the geometry of spacetime}. Here, the semantic content of a token $\mathbf{x}_i$ acts as ``matter'' that dynamically curves and scales the local manifold metric (Fig.~\ref{fig:gravitational_lens}):
\begin{eqbox}
\begin{equation}
    s_i = \sigma(\Proj(\mathbf{x}_i)),
    \label{eq:metric_gating}
\end{equation}
\end{eqbox}
where $\Proj(\mathbf{x}_i)$ projects the $D$-dimensional token embedding into a scalar ``mass'' representation for each of the $n_h$ attention heads. 

To control the average gating scale across deep residual layers, we instantiate $\sigma$ as the following positive, head-wise \textbf{Mean-Normalized Conformal Parameterization}:
\begin{eqbox}
\begin{equation}
    \tilde{s}_i = \exp\big(\tau \cdot \tanh(\mathbf{x}_i \mathbf{U})\big), \qquad s_i = \frac{\tilde{s}_i + \epsilon}{\frac{1}{N}\sum_{j=1}^N \tilde{s}_j + \epsilon},
    \label{eq:exp_metric}
\end{equation}
\end{eqbox}
where $\mathbf{U} \in \R^{D \times n_h}$ is zero-initialized so that $s_i \equiv 1$ at the start of training, $\tau \in \R^{n_h}$ is a learnable temperature parameter, and $\epsilon>0$ stabilizes the normalization. This parameterization satisfies $\frac{1}{N}\sum_i s_i=1$ in each head. Furthermore, by matrix associativity, $(\mathbf{L}\mathbf{L}^\top \odot \mathbf{C})\mathbf{V} = \diag(\mathbf{L})\big(\mathbf{C}(\diag(\mathbf{L})\mathbf{V})\big)$, avoiding any $\mathcal{O}(N^2)$ outer-product allocation.

\paragraph{Physical Interpretation (Mean-Normalized Gravitational Lensing).}
Equations~\eqref{eq:metric_gating}--\eqref{eq:exp_metric} endow self-attention with profound geometric interpretability. Normalizing $s_i$ to unit mean fixes the average gating scale: allocating larger scales to salient foreground tokens ($s_i > 1$, ``gravity wells'') balances smaller scales elsewhere ($s_j < 1$). Consequently, the outer-product metric $\mathbf{g}_{ij} = s_i s_j$ spatially amplifies foreground feature interactions while compressing background noise, acting as a \textbf{``Gravitational Lens''} for information flow.

\section{Inducing Gauge Structure on the Value Bundle}
\label{sec:value_bundle}

Formulating value vectors $\{\mathbf{v}_i\}$ as sections of a vector bundle $\mathcal{E}_V = \mathcal{M} \times \R^{D_v}$ decoupled from the tangent bundle $T\mathcal{M}$ allows us to introduce independent geometric operations during feature aggregation. Let $C_{m,i}$ denote the attention weight between query $m$ and key $i$ from Eq.~\eqref{eq:gaugeattn}. We formulate three structured extensions on $\mathcal{E}_V$:

\begin{enumerate}[label=\textbf{\arabic*.}]
    \item \textbf{Dynamic Fiber Gating:} Rather than scaling all channels of $\mathbf{v}_i$ uniformly by the scalar $C_{m,i}$, we define a query-conditioned fiber map $g: T_{\mathbf{p}_m}\mathcal{M} \to \R^{D_v}$ for content-addressable channel selection: $\mathbf{o}_m = \sum_i C_{m,i} \left(g(\mathbf{q}_m) \odot \mathbf{v}_i\right)$.
    \item \textbf{Local Gauge Symmetry via Group Action:} For data with local orientation symmetries, we learn a section $h: \mathcal{M} \to G$ mapping each position to a Lie group $G \subset \SO(D_v)$ acting on the value fiber: $\mathbf{o}_m = \sum_i C_{m,i} \left(h(\mathbf{p}_i)\cdot \mathbf{v}_i\right)$.
    \item \textbf{Independent Connections on the Value Bundle:} Challenging the global uniformity of a trivial bundle, we introduce a learnable value connection $\mathcal{C}$ with parallel transport operator $\mathcal{C}_{i\to m}: \mathcal{E}_{V,\mathbf{p}_i} \to \mathcal{E}_{V,\mathbf{p}_m}$, yielding:
    \begin{eqbox}
    \begin{equation}
        \mathbf{o}_m = \sum_{i=1}^N C_{m,i}\, \mathcal{C}_{i\to m}(\mathbf{v}_i).
    \end{equation}
    \end{eqbox}
\end{enumerate}

\paragraph{Concrete Realization: The Locality Focusing Mechanism.}
This dual-geometry framework—where the tangent bundle connection governs \emph{where to attend} ($Q/K$) and the value connection $\mathcal{C}_{i\to m}$ governs \emph{how to integrate} ($V$)—admits a highly effective, minimal-overhead instantiation for visual tasks, which we term \textbf{Locality Focusing (LF)}.

Vanilla Vision Transformers lack the intrinsic locality bias of CNNs and therefore struggle on small datasets. We instantiate the value transport operator $\mathcal{C}_{i\to m}$ as a spatial attenuation connection $\mathcal{C}_{i\to m}(\mathbf{v}_i) = \lambda_{m,i} \mathbf{v}_i$, where $\lambda_{m,i} \in (0, 1]$ decays monotonically with the geodesic distance between positions $\mathbf{d}_m$ and $\mathbf{d}_i$:
\begin{eqbox}
\begin{equation}
    \mathbf{o}_m = \sum_{n=1}^N \underbrace{C_{m,n}(\mathbf{q}_m, \mathbf{k}_n)}_{\text{Semantic / Gauge Similarity}} \underbrace{G_\sigma(\mathbf{d}_m, \mathbf{d}_n)}_{\text{Spatial Proximity ($\lambda_{m,n}$)}} \mathbf{v}_n, \quad \text{with} \quad G_\sigma(\mathbf{x}, \mathbf{y}) = \exp\left(-\frac{\|\mathbf{x}-\mathbf{y}\|_{\mathbf{A}}^2}{2\sigma^2}\right),
    \label{eq:bilateral_filter}
\end{equation}
\end{eqbox}
where $\mathbf{A} \succ 0$ and $\sigma$ are learnable parameters (initialized progressively from sharper local bandwidths in early layers to broader receptive fields in deeper layers). Strikingly, Eq.~\eqref{eq:bilateral_filter} mirrors the classic \textbf{bilateral filter} in image processing, disentangling semantic intensity similarity from spatial domain proximity. Collecting $\Lambda_{m,n} = \lambda_{m,n}$ into a spatial kernel $\mathbf{\Lambda} \in \R^{N \times N}$, we apply row-wise bilateral normalization so that spatial focusing sharpens neighborhood selectivity while preserving total attention energy:
\begin{eqbox}
\begin{equation}
    \widetilde{\mathbf{C}} = \mathrm{RowNorm}(\mathbf{C} \odot \mathbf{\Lambda}), \qquad \mathbf{O} = \diag(\mathbf{L})\left(\widetilde{\mathbf{C}}\big(\diag(\mathbf{L})\mathbf{V}\big)\right).
    \label{eq:bilateral_norm}
\end{equation}
\end{eqbox}

\section{Engineering Practicality: Translation Symmetries, FlashAttention, and Linear Complexity}
\label{sec:engineering}

While differential geometry and gauge theory provide a rigorous foundation for modeling curved attention manifolds, practical deployment in modern large-scale Vision and Language Models motivates three engineering objectives: (1) \textbf{translation symmetries}, (2) compatibility with IO-aware fused kernels (\textbf{FlashAttention}), and (3) scalability to long sequences via \textbf{$\mathcal{O}(N)$ linear complexity}. We establish the positional symmetry of each curvature form and derive efficient evaluation schemes through gauge factorization and systematic linearization.

\subsection{Relative Positional Invariance and Magnetic Translation Covariance}

A hallmark of modern positional encodings like RoPE is translation invariance: shifting all token coordinates by a global offset $\mathbf{c}$ ($\mathbf{p}_i \mapsto \mathbf{p}_i + \mathbf{c}$) leaves attention scores unchanged. In our factorized transport operator $\mathcal{P}_{n\to m} = \exp(\Delta \mathbf{p}_{nm}\cdot \mathbf{J}_L)\exp(\lambda \mathit{\Omega}_{nm}\cdot \mathbf{J}_R)$ (Eq.~\eqref{eq:factorized_transport}), the left-isoclinic component depends explicitly on the relative displacement $\Delta \mathbf{p}_{nm} = \mathbf{p}_m - \mathbf{p}_n$. 

For the right-isoclinic phase $\mathit{\Omega}_{nm}$, the curvature parameterization determines the translation symmetry:
\begin{enumerate}[label=\arabic*)]
    \item \textbf{Strict Relative Invariance:} For orthogonal spatial coupling $\mathit{\Omega}_{nm}^{\mathrm{ortho}}=(y_m-y_n)t\theta_B$ (Eq.~\eqref{eq:omega_ortho}), the transport depends only on coordinate differences. The content-induced phases in Eq.~\eqref{eq:omega_dyn} are independent of the coordinate origin. These choices preserve attention scores under global coordinate shifts with token content held fixed.
    \item \textbf{Magnetic Translation Covariance (for Symplectic Area):} Even for the coordinate wedge product $\mathit{\Omega}_{nm}^{\mathrm{symp}} = \mathbf{p}_n \wedge \mathbf{p}_m = x_n y_m - y_n x_m$ (Eq.~\eqref{eq:omega_symp}), shifting the origin by $\mathbf{c} = (c_x, c_y)$ transforms the curvature form as:
    \begin{eqbox}
    \begin{equation}
        \mathit{\Omega}_{nm}' = (\mathbf{p}_n + \mathbf{c}) \wedge (\mathbf{p}_m + \mathbf{c}) = \mathbf{p}_n \wedge \mathbf{p}_m + \underbrace{(\mathbf{c}\wedge \mathbf{p}_m)}_{\alpha_m} - \underbrace{(\mathbf{c}\wedge \mathbf{p}_n)}_{\alpha_n} = \mathit{\Omega}_{nm} + \alpha_m - \alpha_n.
        \label{eq:magnetic_translation}
    \end{equation}
    \end{eqbox}
    The shift-induced discrepancy $\alpha_m-\alpha_n$ is a separable pure-gauge phase. Define $\mathbf{G}_i=\exp(\lambda\alpha_i\mathbf{J}_R)$. Chiral commutativity gives the endpoint transformation law
    \begin{eqbox}
    \begin{equation}
        \mathcal{P}'_{n\to m}=\mathbf{G}_m\mathcal{P}_{n\to m}\mathbf{G}_n^{-1},\qquad
        (\mathbf{G}_m\mathbf{q}_m)^\top\mathcal{P}'_{n\to m}(\mathbf{G}_n\mathbf{k}_n)
        =\mathbf{q}_m^\top\mathcal{P}_{n\to m}\mathbf{k}_n.
        \label{eq:magnetic_covariance}
    \end{equation}
    \end{eqbox}
    Thus, magnetic translation preserves the score under the joint transformation of coordinates and endpoint features. With fixed Q/K, the score is coordinate-dependent, while the closed-loop phase remains strictly shift-invariant:
    \begin{eqbox}
    \begin{equation}
        \mathit{\Omega}_{nm}' + \mathit{\Omega}_{ml}' + \mathit{\Omega}_{ln}' = (\mathbf{p}_m - \mathbf{p}_n) \wedge (\mathbf{p}_l - \mathbf{p}_m) = 2\,\mathrm{Area}(\triangle nml).
    \end{equation}
    \end{eqbox}
\end{enumerate}

\subsection{Weak-Field Linearization for Native FlashAttention}

Although Theorem~\ref{thm:fast_holonomy} reduces the exact evaluation of non-separable curvature $\mathit{\Omega}_{nm}$ from $\mathcal{O}(N^2 D^2)$ to $\mathcal{O}(N^2 D)$, multiplying the $N \times N$ score matrix by $\cos(\lambda \mathit{\Omega}_{nm})$ and $\sin(\lambda \mathit{\Omega}_{nm})$ still requires materializing the attention matrix in high-bandwidth memory (HBM), precluding standard fused \textbf{FlashAttention} kernels when $\mathit{\Omega}_{nm}$ is non-separable.

To construct token-wise features for off-the-shelf FlashAttention, we invoke the \textbf{weak-field approximation} ($\max_{m,n}|\lambda\mathit{\Omega}_{nm}|\ll 1$) from gauge physics, linearizing the right-chiral curvature exponential in the Lie algebra $\su(2)_R$:
\begin{eqbox}
\begin{equation}
    \exp(\lambda \mathit{\Omega}_{nm} \mathbf{J}_R) = \mathbf{I} + \lambda \mathit{\Omega}_{nm} \mathbf{J}_R + \mathcal{O}(\lambda^2).
    \label{eq:weak_field}
\end{equation}
\end{eqbox}

\begin{proposition}[Token-Wise Decoupling under Weak-Field Linearization]
\label{prop:flash_linearization}
Let $\tilde{\mathbf{q}}_m = \exp(-\mathbf{p}_m \cdot \mathbf{J}_L)\mathbf{q}_m$ and $\tilde{\mathbf{k}}_n = \exp(-\mathbf{p}_n \cdot \mathbf{J}_L)\mathbf{k}_n$ carry the exact $\SO(4)$ relative translation. Under first-order weak-field linearization, both the symmetric additive phase $\mathit{\Omega}_{nm}^{(+)} = \frac{1}{2}(\omega_m + \omega_n)$ and rank-2 wedge curvature $\mathit{\Omega}_{nm}^{(\wedge)} = \phi_{1,n}\phi_{2,m} - \phi_{2,n}\phi_{1,m}$ factorize into standard token-wise inner products $\langle \hat{\mathbf{q}}_m, \hat{\mathbf{k}}_n \rangle$:
\begin{enumerate}[label=(\roman*)]
    \item \textbf{In-Place Tangent Perturbation ($D$-dimensional):} Defining token-wise perturbed vectors
    \begin{eqbox}
    \begin{equation}
        \hat{\mathbf{q}}_m = \left(\mathbf{I} - \frac{\lambda}{2}\omega(\mathbf{x}_m)\mathbf{J}_R\right)\tilde{\mathbf{q}}_m, \qquad \hat{\mathbf{k}}_n = \left(\mathbf{I} + \frac{\lambda}{2}\omega(\mathbf{x}_n)\mathbf{J}_R\right)\tilde{\mathbf{k}}_n,
        \label{eq:flash_inplace}
    \end{equation}
    \end{eqbox}
    and using $\mathbf{J}_R^\top = -\mathbf{J}_R$ and $\mathbf{J}_R^2 = -\mathbf{I}$, their standard Euclidean dot product yields:
    \begin{eqbox}
    \begin{equation}
        \hat{\mathbf{q}}_m^\top \hat{\mathbf{k}}_n = \tilde{\mathbf{q}}_m^\top \left( \left(1 - \frac{\lambda^2}{4}\omega_m\omega_n\right)\mathbf{I} + \frac{\lambda}{2}(\omega_m + \omega_n)\mathbf{J}_R \right) \tilde{\mathbf{k}}_n \approx \tilde{\mathbf{q}}_m^\top \exp\left(\frac{\lambda}{2}(\omega_m + \omega_n)\mathbf{J}_R\right)\tilde{\mathbf{k}}_n.
    \end{equation}
    \end{eqbox}
    \item \textbf{Exact Bilinear Wedge Decomposition ($3D$-dimensional):} For $\mathit{\Omega}_{nm}=x_n y_m-y_n x_m$, define concatenated token-wise features
    \begin{eqbox}
    \begin{equation}
        \hat{\mathbf{q}}_m=\begin{bmatrix}\tilde{\mathbf{q}}_m\\y_m\tilde{\mathbf{q}}_m\\x_m\tilde{\mathbf{q}}_m\end{bmatrix},\qquad
        \hat{\mathbf{k}}_n=\begin{bmatrix}\tilde{\mathbf{k}}_n\\\lambda x_n\mathbf{J}_R\tilde{\mathbf{k}}_n\\-\lambda y_n\mathbf{J}_R\tilde{\mathbf{k}}_n\end{bmatrix}.
        \label{eq:flash_wedge_features}
    \end{equation}
    \end{eqbox}
    Their dot product is exactly $\tilde{\mathbf{q}}_m^\top[\mathbf{I}+\lambda(x_n y_m-y_n x_m)\mathbf{J}_R]\tilde{\mathbf{k}}_n$. The same construction applies to the dynamic wedge by replacing $(x_i,y_i)$ with $(\phi_{1,i},\phi_{2,i})$. This feature lift evaluates the first-order score with a constant-factor channel expansion and no $N\times N$ intermediate matrix.
\end{enumerate}
Furthermore, absorbing the row-wise query metric scale $s_m$ to the output of attention and folding the key scale $s_n$ into the value vector $\hat{\mathbf{v}}_n = s_n \mathbf{v}_n$ evaluates the resulting weak-field \textrm{GaugeAttn} operator with a single native FlashAttention call:
\begin{eqbox}
\begin{equation}
    \mathrm{GaugeAttn}_{\mathrm{Flash}}(\mathbf{Q}, \mathbf{K}, \mathbf{V})_m = s_m \cdot \mathrm{FlashAttention}\big(\hat{\mathbf{Q}}, \hat{\mathbf{K}}, \diag(\mathbf{L})\mathbf{V};\,\mathrm{scale}=d_k^{-1/2}\big)_m.
    \label{eq:flash_gaugeattn}
\end{equation}
\end{eqbox}
\end{proposition}
Equation~\eqref{eq:flash_gaugeattn} is remarkably practical: all gauge rotations ($\mathbf{J}_L, \mathbf{J}_R$) and metric scalings ($s_m, s_n$) are executed as lightweight $\mathcal{O}(ND)$ element-wise/block-wise pre- and post-processing steps around a native fused attention call, achieving $\mathcal{O}(N)$ HBM memory footprint for fixed feature dimension. The wedge construction uses $3d_k$ query/key channels, and the explicit scale $d_k^{-1/2}$ retains the original attention temperature. For kernels with equal Q/K/V head widths, zero-padding the values to the lifted width and selecting the original value channels after attention preserves the output exactly.

\paragraph{Canonical Chiral Basis for Split-Channel Execution.}
To further reduce rotational FLOPs in latency-critical applications, we can specialize the general chiral generators $\mathbf{J}_L, \mathbf{J}_R \in \so(4)$ by fixing both unit rotation axes to the canonical $z$-axis, $\mathbf{a} = \mathbf{b} = (0, 0, 1)^\top$. In this canonical basis, $\mathbf{J}_L=A_3$ and $\mathbf{J}_R=B_3$ act within two channel planes. Both planes use the block-wise rotation generator:
\begin{eqbox}
\begin{equation}
    \mathbf{J}_{LR} \triangleq A_3 = \begin{bmatrix} 0 & -1 & 0 & 0 \\ 1 & 0 & 0 & 0 \\ 0 & 0 & 0 & -1 \\ 0 & 0 & 1 & 0 \end{bmatrix} = \diag(J_2, J_2), \quad \text{where} \quad J_2 = \begin{bmatrix} 0 & -1 \\ 1 & 0 \end{bmatrix}.
    \label{eq:j_lr_split}
\end{equation}
\end{eqbox}
The canonical generators satisfy $A_3=\diag(J_2,J_2)$ and $B_3=\diag(J_2,-J_2)$. Hence, with $R_2(\eta)=\exp(\eta J_2)$, the original chiral phases obey
\begin{eqbox}
\begin{equation}
    \exp(\theta A_3)\exp(\beta B_3)
    =\diag\big(R_2(\theta+\beta),R_2(\theta-\beta)\big).
    \label{eq:canonical_chiral_angles}
\end{equation}
\end{eqbox}
We parameterize the split-channel variant directly by its two plane angles $\eta_{\mathrm{pos}}=\Delta p_{nm}$ and $\eta_{\mathrm{curv}}=\lambda\mathit{\Omega}_{nm}$, where $\Delta p_{nm}=\Delta\mathbf{p}_{nm}^\top\boldsymbol{\omega}_L$. This corresponds to the chiral reparameterization $\theta=(\eta_{\mathrm{pos}}+\eta_{\mathrm{curv}})/2$ and $\beta=(\eta_{\mathrm{pos}}-\eta_{\mathrm{curv}})/2$. It assigns $\R^4=\R^2_{\mathrm{pos}}\oplus\R^2_{\mathrm{curv}}$ to position and curvature through
\begin{eqbox}
\begin{equation}
    \mathcal{P}_{n\to m}^{\mathrm{split}} = \exp\begin{bmatrix} \Delta p_{nm} J_2 & \mathbf{0} \\ \mathbf{0} & \lambda \mathit{\Omega}_{nm} J_2 \end{bmatrix} = \begin{bmatrix} \cos(\Delta p_{nm})I_2 + \sin(\Delta p_{nm})J_2 & \mathbf{0} \\ \mathbf{0} & \cos(\lambda \mathit{\Omega}_{nm})I_2 + \sin(\lambda \mathit{\Omega}_{nm})J_2 \end{bmatrix}.
    \label{eq:split_transport}
\end{equation}
\end{eqbox}
The split-channel parameterization implements each plane with the standard two-channel rotation primitive used in RoPE~\cite{su2024roformer}. For separable phases $\Delta p_{nm}=f_m-f_n$ and $\lambda\mathit{\Omega}_{nm}=h_m-h_n$, define $\mathbf{R}^{\mathrm{split}}_i=\diag(R_2(f_i),R_2(h_i))$; then $\mathcal{P}^{\mathrm{split}}_{n\to m}=\mathbf{R}^{\mathrm{split}}_m(\mathbf{R}^{\mathrm{split}}_n)^\top$. Token-wise rotations by $(\mathbf{R}^{\mathrm{split}}_i)^\top$ therefore provide an exact fused FlashAttention implementation. Non-separable phases use pairwise evaluation or the weak-field feature lift in Proposition~\ref{prop:flash_linearization}.

\subsection{$\mathcal{O}(N)$ Linear GaugeAttention}

For ultra-long sequences ($N \gg D$) or high-resolution visual streams where even $\mathcal{O}(N^2)$ time complexity becomes a bottleneck, we can further simplify our framework into a strict \textbf{$\mathcal{O}(N D^2)$ Linear GaugeAttention} architecture.

A notorious weakness of conventional Linear Attention which replaces softmax with a kernel feature map $\phi(\mathbf{q}_m)^\top \phi(\mathbf{k}_n)$—is \emph{memory dilution}: because all tokens write into the global KV memory state $\sum_n \phi(\mathbf{k}_n)\mathbf{v}_n^\top$ with uniform unit weight, background clutter rapidly overwhelms salient signals.

Our Dynamic Metric Gating $\mathbf{g}_{mn} = s_m s_n$ and orthogonal $\SO(4)$ gauge transport resolve this limitation naturally. Specifically, let $\phi(\mathbf{u}) = \mathrm{Normalize}(\mathrm{elu}(\mathbf{u})+\mathbf{1})$ be a spherical non-negative kernel projection, and let $\mathbf{R}_i = \exp(x_i t\theta_L\mathbf{J}_L + y_i t\theta_B\mathbf{J}_R) \in \SO(D)$ denote the exact chiral gauge frame at position $i$. Defining the gauge-rotated kernel features $\hat{\mathbf{q}}_m = \mathbf{R}_m^\top \phi(\mathbf{q}_m)$ and $\hat{\mathbf{k}}_n = \mathbf{R}_n^\top \phi(\mathbf{k}_n)$, their inner product preserves exact $\SO(4)$ relative positional invariance and bounded norm $\hat{\mathbf{q}}_m^\top \hat{\mathbf{k}}_n \in [-1, 1]$:
\begin{eqbox}
\begin{equation}
    \hat{\mathbf{q}}_m^\top \hat{\mathbf{k}}_n = \phi(\mathbf{q}_m)^\top \mathbf{R}_m \mathbf{R}_n^\top \phi(\mathbf{k}_n) = \phi(\mathbf{q}_m)^\top \mathcal{P}_{n\to m}(\Delta \mathbf{p}_{nm}) \phi(\mathbf{k}_n).
\end{equation}
\end{eqbox}

Exploiting the rank-1 multiplicative separability of our conformal metric $\mathbf{g}_{mn} = s_m s_n$ and matrix multiplication associativity, the Linear GaugeAttention output $\mathbf{o}_m$ collapses from $\mathcal{O}(N^2 D)$ to $\mathcal{O}(N D D_v)$:
\begin{eqbox}
\begin{equation}
    \mathbf{o}_m = \frac{1}{\sqrt{N}}\sum_{n=1}^N \overbrace{(s_m s_n)}^{\text{Metric }\mathbf{g}_{mn}} \overbrace{\left(\hat{\mathbf{q}}_m^\top \hat{\mathbf{k}}_n\right)}^{\text{Relative }\SO(4)\text{ Gauge}} \mathbf{v}_n = s_m \hat{\mathbf{q}}_m^\top \underbrace{\left( \frac{1}{\sqrt{N}}\sum_{n=1}^N \big(\overbrace{s_n}^{\text{Gravitational Mass}} \hat{\mathbf{k}}_n\big) \mathbf{v}_n^\top \right)}_{\text{Gravitational Memory State }\mathbf{M}_{\mathrm{grav}} \in \R^{D \times D_v}} = s_m \hat{\mathbf{q}}_m^\top \mathbf{M}_{\mathrm{grav}},
    \label{eq:linear_gaugeattn}
\end{equation}
\end{eqbox}
where the \textbf{Gravitational Memory State} $\mathbf{M}_{\mathrm{grav}} \in \R^{D \times D_v}$ is computed once globally in $\mathcal{O}(N D D_v)$ time (or normalized by $\phi(\mathbf{q}_m)^\top \sum_n s_n \phi(\mathbf{k}_n)$ when unnormalized kernels are used). Here, the dynamic scale $s_n = \sigma(\Proj(\mathbf{x}_n))$ plays a vital physical role: tokens with low semantic mass ($s_n \to 0$) are prevented from polluting the global memory matrix $\mathbf{M}_{\mathrm{grav}}$, while salient foreground tokens ($s_n > 1$) dominate memory write-in. Moreover, if the value-bundle Locality Focusing connection $\lambda_{m,n} = \gamma^{m-n}$ ($m \ge n$) is incorporated into causal sequence modeling, Eq.~\eqref{eq:linear_gaugeattn} directly becomes a gated recurrent state space update $\mathbf{M}_m = \gamma \mathbf{M}_{m-1} + s_m \hat{\mathbf{k}}_m \mathbf{v}_m^\top$, bridging differential geometry with modern linear RNNs and State Space Models.

\begin{remark}[\textbf{Native Compatibility with Dynamic Sparse Attention and Token Pruning}]
Beyond dense and linear regimes, the rank-1 factorized metric $\mathbf{g}_{mn} = s_m s_n$ naturally serves as a zero-overhead importance router for \emph{Dynamic Sparse Attention}. Because the ``gravitational mass'' $s_n = \sigma(\Proj(\mathbf{x}_n))$ is evaluated in $\mathcal{O}(N)$ time prior to pairwise $\mathbf{Q}\mathbf{K}^\top$ computation, tokens with negligible mass ($s_n < \tau_{\mathrm{grav}}$) can be pruned directly from the key--value cache. Combined with the spatial decay of Locality Focusing ($\lambda_{m,n}$), this induces a physically grounded sparse topology where low-mass background tokens interact solely within local geodesic windows, whereas high-mass foreground tokens ($s_n \gg 1$) act as global gravitational beacons with long-range receptive fields.
\end{remark}

\begin{table}[th]
\centering
\caption{\textbf{Complexity Spectrum of GaugeAttention Variants.} Our framework spans exact abelian holonomy in $\SO(4)$, token-wise FlashAttention formulations, and $\mathcal{O}(N)$ Linear Attention with mean-normalized metric gating. Positional symmetry follows the phase parameterization.}
\label{tab:complexity_spectrum}
\vspace{0.3em}
\renewcommand{\arraystretch}{1.25}
\setlength{\tabcolsep}{4pt}
\resizebox{\textwidth}{!}{%
\begin{tabular}{l | c c c c c l}
\toprule
\textbf{Model Variant} & \textbf{Curvature $\mathit{\Omega}_{nm}$} & \textbf{Positional Symmetry} & \textbf{FlashAttn Ready} & \textbf{Time} & \textbf{Memory} & \textbf{Target Regime} \\
\midrule
\textbf{1. Exact GaugeAttn} (Thm.~\ref{thm:fast_holonomy}) & Exact $\exp(\lambda \mathit{\Omega}_{nm}\mathbf{J}_R)$ & Magnetic covariance (symplectic) & $\times$ ($2\times$ GEMM) & $\mathcal{O}(N^2 D)$ & $\mathcal{O}(N^2)$ & Dense 2D/3D Vision \\
\textbf{2. Flash GaugeAttn} (Prop.~\ref{prop:flash_linearization}) & Separable / Weak-Field $\mathcal{O}(\lambda)$ & Shift invariant (separable/content) & \textbf{$\checkmark$ (Native Fused)} & $\mathcal{O}(N^2 D)$ & \textbf{$\mathcal{O}(N)$} & ViT \& LLM Training \\
\textbf{3. Linear GaugeAttn} (Eq.~\ref{eq:linear_gaugeattn}) & Orthogonal $\SO(4)$ Kernel & Shift invariant & \textbf{$\checkmark$ (Linear Scan)} & \textbf{$\mathcal{O}(N D^2)$} & \textbf{$\mathcal{O}(D^2)$} & Long Context ($N \gg D$) \\
\bottomrule
\end{tabular}%
}
\end{table}

Table~\ref{tab:complexity_spectrum} summarizes the resulting three-tier complexity spectrum of GaugeAttention. Depending on the computational budget and sequence length $N$, practitioners can seamlessly transition between exact non-abelian holonomy (\textbf{Exact GaugeAttn}, $\mathcal{O}(N^2 D)$ time), IO-aware hardware-fused execution (\textbf{Flash GaugeAttn}, $\mathcal{O}(N)$ HBM memory), and strict linear-complexity state aggregation (\textbf{Linear GaugeAttn}, $\mathcal{O}(N D^2)$ time and $\mathcal{O}(D^2)$ memory), all while preserving relative positional invariance and dynamic gravitational metric gating.

\section{Experiments}
\label{sec:experiments}

We evaluate GaugeAttention on CIFAR-100, which contains $50{,}000$ training images and $10{,}000$ test images across $100$ classes. Using a shared Conv-Stem ViT backbone, we compare relative rotations, spatial and content-dependent curvature phases, dynamic metric gating, and Locality Focusing against standard self-attention. We also evaluate split-channel FlashAttention and kernelized Linear GaugeAttention on the same task.

\subsection{Experimental Architecture and Setup}

\paragraph{Minimal Conv-Stem ViT backbone.}
Two $3 \times 3$ stride-$2$ convolutions with BatchNorm and GELU~\cite{xiao2021early} map each $32 \times 32 \times 3$ image through $96$ intermediate channels to an $8 \times 8$ grid of $N=64$ tokens with embedding dimension $D=192$. The stem has a $7 \times 7$ receptive field and $169{,}056$ learnable parameters. Four Transformer layers with $6$ heads, head dimension $d_k=32$, and MLP expansion ratio $3$ process the tokens, followed by LayerNorm, global average pooling, and a linear classifier. In the chiral $\SO(4)$ variants, the $K=8$ four-dimensional blocks in each head share the rotation axes.

The variants have approximately $1.67$--$1.69$ million learnable parameters. In the full configuration, the spatial-wedge and content-wedge variants have $1{,}677{,}436$ and $1{,}686{,}652$ parameters, respectively. The content projection adds $9{,}216$ parameters across the four layers.

\paragraph{Training protocol.}
We train all models from scratch for $200$ epochs on a single NVIDIA RTX 4090 GPU using PyTorch 2.7. We use AdamW with batch size $128$, initial learning rate $10^{-3}$, cosine learning rate decay, and weight decay $0.05$. Data augmentation consists of random cropping, horizontal flipping, AutoAugment, and Random Erasing. Each variant is trained independently with seeds $42$, $43$, and $44$. We report the arithmetic mean of the three CIFAR-100 test top-1 accuracies.

\paragraph{Variant naming.}
The \texttt{\_base} suffix denotes variants with both dynamic metric gating and Locality Focusing (LF) disabled. The \texttt{\_gravity} suffix enables only the dynamic metric gate $s_m s_n$ from Eq.~\eqref{eq:gaugeattn}, with scales defined by Eq.~\eqref{eq:exp_metric}. The \texttt{\_full} suffix enables both the metric gate and LF from Eq.~\eqref{eq:bilateral_norm}. Disabling the gate sets $s_i=1$ for every token, including in the linear formulation of Eq.~\eqref{eq:linear_gaugeattn}. The Euclidean and rotational baselines also disable both components. In particular, \texttt{so4\_rot} provides the ungated baseline for the orthogonal-phase variants.

\begin{table}[t!]
\centering
\caption{\textbf{CIFAR-100 test accuracy with a shared Conv-Stem ViT.} Top-1 accuracy is averaged over seeds $42$, $43$, and $44$. Gains are in percentage points relative to \texttt{vanilla\_ape}, which uses learned absolute positional embeddings. Complexity refers to the attention computation.}
\label{tab:convstem_results}
\vspace{0.3em}
\renewcommand{\arraystretch}{1.25}
\setlength{\tabcolsep}{4.2pt}
\resizebox{\textwidth}{!}{%
\begin{tabular}{l | c c c c c | c c}
\toprule
\textbf{Model and equation} & \textbf{Rotation} & \textbf{Position or curvature term} & \textbf{Metric gate} & \textbf{LF} & \textbf{Complexity} & \textbf{Top-1 (\%)} & \textbf{Gain (pp)} \\
\midrule
\multicolumn{8}{l}{\textit{\textbf{Euclidean baselines}}} \\
\texttt{vanilla\_nope} \enspace Eq.~\eqref{eq:vanilla_attn} & None & None & Fixed & -- & $\mathcal{O}(N^2D)$ & 60.98 & $-4.53$ \\
\texttt{vanilla\_ape} \enspace Eq.~\eqref{eq:vanilla_attn} & None & Learned APE & Fixed & -- & $\mathcal{O}(N^2D)$ & 65.51 & $0.00$ \\
\midrule
\multicolumn{8}{l}{\textit{\textbf{Ungated rotational baselines}}} \\
\texttt{so2\_rot} \enspace Eq.~\eqref{eq:split_transport} & 2D RoPE & Separable $(\Delta x_{nm},\Delta y_{nm})$ & Fixed & -- & $\mathcal{O}(N^2D)$ & 67.92 & $+2.41$ \\
\texttt{so4\_rot} \enspace Eq.~\eqref{eq:omega_ortho} & Chiral $\SO(4)$ & Separable $(\Delta x_{nm},\Delta y_{nm})$ & Fixed & -- & $\mathcal{O}(N^2D)$ & 68.66 & $+3.15$ \\
\midrule
\multicolumn{8}{l}{\textit{\textbf{Linear and split-channel attention}}} \\
\texttt{so4\_linear\_base} \enspace Eq.~\eqref{eq:linear_gaugeattn} & Chiral $\SO(4)$ & Separable $(\Delta x_{nm},\Delta y_{nm})$ & Fixed & -- & $\mathcal{O}(ND^2)$ & 66.04 & $+0.53$ \\
\texttt{so4\_linear\_gravity} \enspace Eq.~\eqref{eq:linear_gaugeattn} & Chiral $\SO(4)$ & Separable $(\Delta x_{nm},\Delta y_{nm})$ & Dynamic & -- & $\mathcal{O}(ND^2)$ & 66.44 & $+0.93$ \\
\texttt{split\_flash\_base} \enspace Eq.~\eqref{eq:split_transport} & $\SO(2)\times\SO(2)$ & Separable split phases & Fixed & -- & $\mathcal{O}(N^2D)$ & 67.93 & $+2.42$ \\
\texttt{split\_flash\_gravity} \enspace Eq.~\eqref{eq:split_transport} & $\SO(2)\times\SO(2)$ & Separable split phases & Dynamic & -- & $\mathcal{O}(N^2D)$ & 68.91 & $+3.40$ \\
\texttt{split\_full} \enspace Eq.~\eqref{eq:split_transport} & $\SO(2)\times\SO(2)$ & Separable split phases & Dynamic & $\checkmark$ & $\mathcal{O}(N^2D)$ & 68.03 & $+2.52$ \\
\midrule
\multicolumn{8}{l}{\textit{\textbf{Chiral $\SO(4)$ with dynamic metric gating}}} \\
\texttt{so4\_ortho\_gravity} \enspace Eq.~\eqref{eq:omega_ortho} & Chiral $\SO(4)$ & Separable $(\Delta x_{nm},\Delta y_{nm})$ & Dynamic & -- & $\mathcal{O}(N^2D)$ & 69.27 & $+3.76$ \\
\texttt{so4\_ortho\_full} \enspace Eq.~\eqref{eq:omega_ortho} & Chiral $\SO(4)$ & Separable $(\Delta x_{nm},\Delta y_{nm})$ & Dynamic & $\checkmark$ & $\mathcal{O}(N^2D)$ & 69.24 & $+3.73$ \\
\texttt{so4\_symplectic\_gravity} \enspace Eq.~\eqref{eq:omega_symp} & Chiral $\SO(4)$ & Spatial wedge $\mathbf{p}_n \wedge \mathbf{p}_m$ & Dynamic & -- & $\mathcal{O}(N^2D)$ & 69.47 & $+3.96$ \\
\texttt{so4\_symplectic\_full} \enspace Eq.~\eqref{eq:omega_symp} & Chiral $\SO(4)$ & Spatial wedge $\mathbf{p}_n \wedge \mathbf{p}_m$ & Dynamic & $\checkmark$ & $\mathcal{O}(N^2D)$ & 69.19 & $+3.68$ \\
\texttt{so4\_dynamic\_gravity} \enspace Eq.~\eqref{eq:omega_dyn} & Chiral $\SO(4)$ & Content wedge $\boldsymbol{\phi}_n \wedge \boldsymbol{\phi}_m$ & Dynamic & -- & $\mathcal{O}(N^2D)$ & 69.45 & $+3.94$ \\
\rowcolor{skyblue!12!white}
\texttt{so4\_dynamic\_full} \enspace Eq.~\eqref{eq:omega_dyn} & Chiral $\SO(4)$ & Content wedge $\boldsymbol{\phi}_n \wedge \boldsymbol{\phi}_m$ & Dynamic & $\checkmark$ & $\mathcal{O}(N^2D)$ & \textbf{69.61} & $\mathbf{+4.10}$ \\
\bottomrule
\end{tabular}%
}
\end{table}

\subsection{Results and Ablation Analysis}

Across the three seeds (42, 43, 44), the highest single-run top-1 accuracies were 70.16\% for \texttt{so4\_dynamic\_full} and 69.97\% for \texttt{so4\_dynamic\_gravity}. Their corresponding mean accuracies were $69.61\%$ and $69.45\%$, respectively.

\paragraph{Positional encoding and rotational transport.}
Learned absolute positional embeddings raise mean top-1 accuracy from $60.98\%$ to $65.51\%$ with standard self-attention (Table~\ref{tab:convstem_results}). Replacing absolute embeddings with separable relative rotations yields $67.92\%$ for 2D RoPE and $68.66\%$ for chiral $\SO(4)$, gains of $2.41$ and $3.15$ percentage points over \texttt{vanilla\_ape}. Both rotational baselines use path-independent transport.

\paragraph{Dynamic metric gating.}
With LF disabled, enabling the metric gate increases mean accuracy from $66.04\%$ to $66.44\%$ for Linear GaugeAttention and from $67.93\%$ to $68.91\%$ for split-channel FlashAttention. The same change raises the separable chiral $\SO(4)$ variant from $68.66\%$ (\texttt{so4\_rot}) to $69.27\%$ (\texttt{so4\_ortho\_gravity}). The gains are $0.40$, $0.98$, and $0.61$ percentage points, respectively.

\paragraph{Curvature phases and Locality Focusing.}
With the metric gate enabled and LF disabled, the separable spatial phase reaches $69.27\%$, while the non-separable spatial and content wedges reach $69.47\%$ and $69.45\%$, respectively. The gravity variant leads in three of the four comparisons with full. Adding LF lowers mean accuracy by $0.03$ percentage points for the separable chiral phase and by $0.28$ for the spatial wedge. In the split-channel comparison, the mean drops from $68.91\%$ to $68.03\%$, and the full variant uses explicit attention scores to apply LF. The content wedge is the exception. With LF enabled, \texttt{so4\_dynamic\_full} achieves the highest mean accuracy of $69.61\%$, gaining $0.16$ percentage points over \texttt{so4\_dynamic\_gravity} and $4.10$ over \texttt{vanilla\_ape}. 

\paragraph{Accuracy and attention complexity.}
The split-channel FlashAttention variants implement the separable rotations in Eq.~\eqref{eq:split_transport} by rotating queries and keys token-wise and retain $\mathcal{O}(N^2D)$ attention time. Linear GaugeAttention evaluates the shared kernel memory in Eq.~\eqref{eq:linear_gaugeattn} in $\mathcal{O}(ND^2)$ time for value width proportional to $D$. At $N=64$, the gated linear variant scores $0.93$ percentage points above the APE baseline and $2.67$ below the gated split-channel FlashAttention variant.

\section{Conclusion}

We have presented \textbf{GaugeAttention}, a theoretical and practical framework that unifies Transformer self-attention with Gauge Field Theory and Riemannian Gravity. By identifying the hidden flatness assumption in standard parallel transport $\mathbf{T}_m \mathbf{T}_n^{-1}$, we introduced a gauge-theoretic ``sandwich'' transport operator $\mathcal{P}_{n\to m} = \mathbf{T}_m \cdot \exp(\mathbf{\Omega}_{nm}) \cdot \mathbf{T}_n^{-1}$ capable of modeling intrinsic manifold curvature and path-dependent holonomy. Leveraging the exceptional chiral isomorphism $\so(4) \cong \su(2)_L \oplus \su(2)_R$, we factorized geodesic transport into commuting flat translation and symplectic curvature components. Crucially, we demonstrated that our geometric formulation does not compromise engineering efficiency: it provides relative positional invariance for separable spatial and content-induced phases, magnetic translation covariance with invariant loop phases for symplectic curvature, and native \textbf{FlashAttention} integration through weak-field feature decomposition or separable split-channel rotations. It scales to ultra-long contexts via an \textbf{$\mathcal{O}(N)$ Linear GaugeAttention} mechanism equipped with a gravity-weighted memory state.

\bibliographystyle{unsrt}
\bibliography{refs}

\end{document}